\documentclass[10pt,journal,compsoc]{IEEEtran}
\ifCLASSOPTIONcompsoc
  \usepackage[nocompress]{cite}
\else
  \usepackage{cite}
\fi
\usepackage{url}
\usepackage{epsfig}
\usepackage{graphicx}
\usepackage{amsmath}
\usepackage{amssymb}
\usepackage{multirow}
\usepackage{colortbl}
\usepackage{color}
\usepackage{threeparttable}
\usepackage{booktabs}
\usepackage[table]{xcolor}
\usepackage{multirow}
\usepackage[export]{adjustbox}
\usepackage{subfloat}
\usepackage{tabularx}
\usepackage{makecell}
\usepackage{caption}
\usepackage{subcaption}
\usepackage[misc]{ifsym}
\usepackage[pagebackref,breaklinks=true,colorlinks,citecolor=blue,urlcolor=blue,linkcolor=blue,bookmarks=false]{hyperref}

\usepackage[capitalize]{cleveref}
\crefname{section}{Sec.}{Secs.}
\Crefname{section}{Section}{Sections}
\Crefname{table}{Table}{Tables}
\crefname{table}{Tab.}{Tabs.}
\usepackage{soul}
\setstcolor{blue}

\begin{document}

\title{Pair then Relation: Pair-Net for \\Panoptic Scene Graph Generation}

\author{ 
  Jinghao Wang$^{\ast}$, Zhengyu Wen$^{\ast}$, Xiangtai Li, Zujin Guo, Jingkang Yang, Ziwei Liu \textsuperscript{\Letter}
\IEEEcompsocitemizethanks{
\IEEEcompsocthanksitem J.~Wang and Z.~Wen contribute equally to this work.
\IEEEcompsocthanksitem J.~Wang, Z.~Wen, X.~Li, Z.~Guo, J.~Yang, Z.~Liu are with the S-Lab, Nanyang Technological University, Singapore. \{jinghao003, zhengyu002, xiangtai.li, gu0008in, jingkang001, ziwei.liu\}@ntu.edu.sg 
}
}

\markboth{IEEE TRANSACTIONS ON PATTERN ANALYSIS AND MACHINE INTELLIGENCE,~Vol.~X, No.~X, X}
{Shell \MakeLowercase{\textit{et al.}}: Bare Advanced Demo of IEEEtran.cls for IEEE Computer Society Journals}

\IEEEtitleabstractindextext{
\begin{abstract}
Panoptic Scene Graph (PSG) is a challenging task in Scene Graph Generation (SGG) that aims to create a more comprehensive scene graph representation using panoptic segmentation instead of boxes. 
Compared to SGG, PSG has several challenging problems: pixel-level segment outputs and full relationship exploration (It also considers thing and stuff relation).
Thus, current PSG methods have limited performance, which hinders downstream tasks or applications. 
This work aims to design a novel and strong baseline for PSG. 
To achieve that, we first conduct an in-depth analysis to identify the bottleneck of the current PSG models, finding that inter-object pair-wise recall is a crucial factor that was ignored by previous PSG methods. 
Based on this and the recent query-based frameworks, we present a novel framework: \textbf{Pair then Relation (Pair-Net)}, which uses a Pair Proposal Network (PPN) to learn and filter sparse pair-wise relationships between subjects and objects. Moreover, we also observed the sparse nature of object pairs for both.
Motivated by this, we design a lightweight Matrix Learner within the PPN, which directly learns pair-wised relationships for pair proposal generation.
Through extensive ablation and analysis, our approach significantly improves upon leveraging the segmenter solid baseline.
Notably, our method achieves over 10\% absolute gains compared to our baseline, PSGFormer. The code of this paper is publicly available at \url{https://github.com/king159/Pair-Net}.
\end{abstract}

}

\maketitle

\IEEEdisplaynontitleabstractindextext

\IEEEpeerreviewmaketitle

\section{Introduction}
\label{sec:intro}

\begin{figure*}[!ht]
    \centering
    \includegraphics[width=1.0\linewidth]{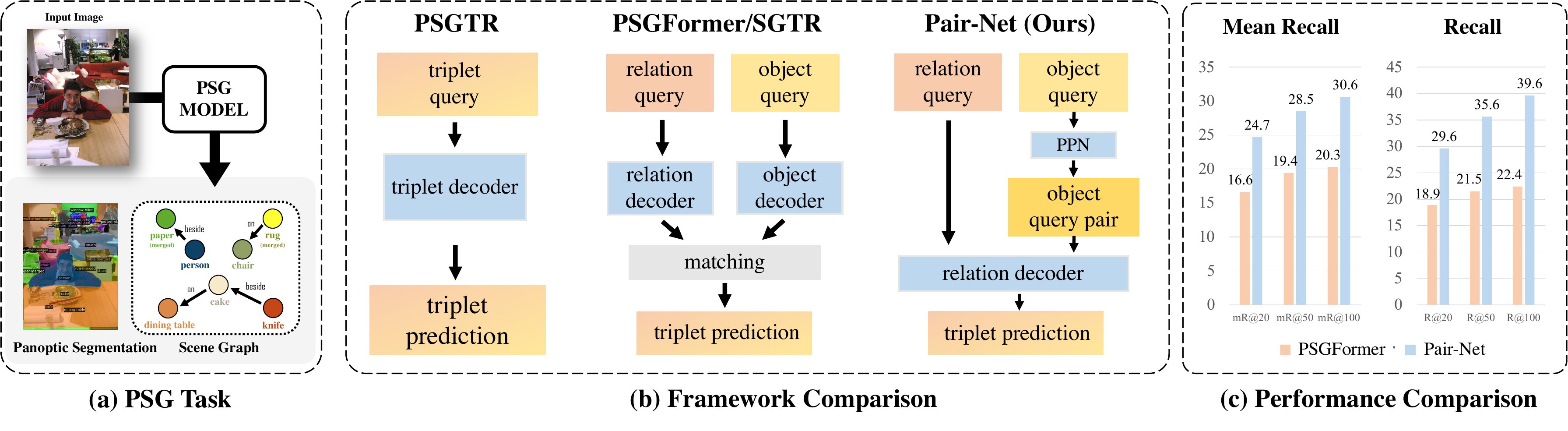}
    \captionof{figure}{\textbf{An illustration of Panoptic Scene Graph (PSG) task, framework, and performance comparisons.} (a) The Panoptic Scene Graph (PSG) task involves generating object-background relations and their masks. (b) Frameworks compared include PSGTR~\cite{psg}, PSGFormer~\cite{psg}, and SGTR~\cite{sgtr}. Our Pair-Net uses the Pair Proposal Network (PPN) to learn object query pairs first and then extract relations between targeted subjects and objects. (c) Performance comparison shows significant improvement over previous methods, demonstrating the effectiveness of Pair-Net.}
    \label{fig:teaser}
    \vspace{-1em}
\end{figure*}

\IEEEPARstart{S}{cene} graph generation (SGG)~\cite{first_sgg} is an essential task in scene understanding that involves generating a graph-structured representation from an input image. This representation captures the locations of a pair of objects (a subject and an object) and their relationship, forming a higher-level abstraction of the image content. SGG has become a fundamental component of several downstream tasks, including image captioning~\cite{captioning1, captioning2, captioning3, captioning4}, visual question answering~\cite{vqa3, vqa4, vqa5}, and visual reasoning~\cite{reasoning1, vg80k}. 

However, current box-based SGG approaches suffer from two primary limitations. \textit{Firstly}, they rely on a coarse object localization provided by a bounding box, which may include noisy foreground pixels belonging to one class. \textit{Secondly}, they do not consider the relationships between background stuff and their context, which is a crucial aspect of scene understanding. To address these limitations, \textbf{Panoptic Scene Graph generation (PSG)} was proposed~\cite{psg}. PSG, as depicted in~\Cref{fig:teaser}(a), leverages a more fine-grained scene mask representation and defines relationships for background stuff, thus offering a more comprehensive understanding of the scene. PSG also provides two one-stage baseline methods, PSGTR and PSGFormer, depicted in \Cref{fig:teaser}(b). Although these methods outperform their two-stage counterparts, their average (triplet) recall rates are only around 10\%. The unsatisfactory performance could fall short of the requirements for downstream applications.

To identify the bottlenecks of the current PSG one-stage models~\cite{psg}, we conducted an in-depth analysis of the calculation of the main recall@K protocol~\cite{psg} of the PSG task. Notice that a successful recall requires a mask-based IOU of over 0.5 for both the subject and object and correct classifications for all elements in the triplet \texttt{\{Subject}, \texttt{Relation}, \texttt{Object\}}, we firstly investigated the segmentation quality of query-based segmenters for isolated subjects/objects to determine their impact on PSG performance. Our experiments demonstrate that a query-based segmenter can recall individual subjects/objects satisfactorily, even without relation training. Therefore, we naturally turn to conjecture that the connectivity of subjects and objects, specifically the recall of subject-object pairing, may affect PSG performance. We obtain evidence for this assumption from experimental results, indicating that the recall of PSG has a strong positive correlation with pair-wise recall, and the absolute pair recall value is far from saturation, suggesting that improving the accuracy of subject-object pairing may be critical for improving PSG performance.
The complete analysis is illustrated in~\Cref{sec:motivation}.

These observations motivate us to propose a new framework for PSG tasks to learn accurate pair-wised relation maps. In this paper, we present Pair-Net, a novel PSG framework, as shown in \Cref{fig:overall}. In Pair-Net, we first apply a query-based segmenter to generate panoptic segmentation for subjects/objects and corresponding object queries without bells and whistles. We then design a Pair-Proposal Network (PPN) that models the object-level interactions between each object, taking the encoded object queries from the segmenter as input and producing feasible subject-object pairs. By systematic analysis of the statistics from the existing scene graph datasets, we notice the strong sparsity of pair-wised relations, which may hinder learning. To acquire sparse and feasible object pairs, we employ a matrix learner to filter the dense pairing relationship map into a considerably sparse one. The frequency count from the ground truth scene graph is used to supervise the output of the matrix learner, significantly improving the sparsity of the filtered map. 
Based on the filtered map, we select Top-K subject-object pairs as inputs to our Relation Fusion module, which predicts the relations from the context information in the given pairs. 
This module utilizes context information from subject-object pairs and facilitates interactions through a cross-attention mechanism. In this way, we eventually generate \texttt{\{Subject}, \texttt{Relation}, \texttt{Object\}} triplets.

We also conduct comprehensive experiments on the PSG dataset. Our method outperforms a strong baseline and achieves a new state-of-the-art performance. In particular, as depicted in~\Cref{fig:teaser} (c), we achieve over 10.2\% improvement compared with PSGFormer~\cite{psg}. Through extensive studies, we demonstrate the effectiveness and efficiency of our proposed model.

In sum, this paper provides the following valuable contributions to the PSG community in the hope of advancing the research in this field:
\begin{enumerate}
    \item \textbf{Comprehensive analysis of pairwise relations in PSG.} We find that although the individual recall for the objects is already saturated for the PSG task, pairwise recall is a significant factor for final recall through systematic experiments.
    \item \textbf{A novel strategy pair-then-relation for solving PSG task.} We explore the pair and then relation generation order and propose a simple but effective PPN for explicit pairing modeling, leading to more precise relationship identification.
    \item \textbf{Significant improvement on all metrics of PSG dataset.} Through extensive experiments, Pair-Net outperforms existing PSG methods by a large margin and achieves new state-of-the-art performance on the PSG dataset.
\end{enumerate}

\section{Related Work}
\label{sec:related_work}

\vspace{2mm}
\noindent
\textbf{Scene Graph Generation.} The existing works for SGG can be divided into the two-stage pipeline and the one-stage pipeline. The two-stage pipeline generally consists of an object detection part and a pairwise predicate estimation part. Many approaches~\cite{imp, motifs, vctree, gpsnet, bi-graph, graphrcnn} model the contextual information between objects. However, these methods are constrained by the high time complexity due to the pairwise predicate estimation, which is infeasible in complex scenarios with many objects but few relations. One-stage pipelines~\cite{fcos, pst, fcn-sgg, structured-rcnn} focus on the one-stage relation detection. However, many still focus on improving detection performance and do not fully use the sparse and semantic priors for SGG. For query-based methods such as RelPN~\cite{relpn} and Graph-RCNN~\cite{graphrcnn}, they generate subject, object, and relation separately and measure triplet compatibility score directly. Also, the ROI-based feature extraction methods make them unable to encode information of stuff. Since the queries given by the segmenter have encoded global semantic information, the relation could be naturally and purely conditioned on the concatenated pair query. Meanwhile, there are also several works for Video Scene Graph Generation ~\cite{teng2021target, video-knet, xu2022meta, gao2022classification, pvsg} and long-tailed problems~\cite{tang2020unbiased, yu2020cogtree, contrastive-loss, logit-adjustment, abdelkarim2021exploring, yang2021probabilistic, balance-adjust, chiou2021recovering, bi-graph, xu2022meta} in SGG. In summary, the methods SGG only handle instance level relation in coarse box format, which lacks pixel-level fine-grained information.

\vspace{2mm}
\noindent
\textbf{Panoptic Scene Graph Generation.} 
The \textit{de facto} SGG dataset, i.e., VG-150~\cite{vg} has some limitations in terms of annotation quality and granularity. 
\textbf{1) Annotation Quality.} VG-150 contains redundant and ambiguous labels of the relation classes (such as \textit{alongside} and \textit{besides}), a large number of trivial triplets (e.g., \textit{woman-has-hair} and \textit{person-has-hand}), and inconsistent and repeated labeling of the same objects (creating many isolated nodes in the scene graph). 
\textbf{2) Annotation Granularity.} A rich and complete visual understanding of an image could not only restrict to only a subset of pixels (i.e., the bounding box used in VG-150) but all pixels of the image~\cite{pq} (i.e., the panoptic mask used in PSG). Also, the interaction (defined as a relation in the scene graph) between background objects (stuff) and foreground objects (thing) is crucial to the understanding of the scene.

To fill the gap and two limitations mentioned above, Panoptic Scene Graph Generation (PSG)~\cite{psg} is proposed. By better defining the relation classes, PSG improves the annotation quality. In addition, by improving the annotation granularity by using all pixels instead of partial of the image, PSG can capture the interaction capture and define a total of 4 kinds of relations: stuff-stuff, thing-stuff, stuff-thing, and thing-thing. Compared with only 1 kind of relation (thing-thing) in SGG, the scene graph in PSG is much more informative and coherent. It improves the difficulty of the SGG task and requires the training model to have a stronger understanding of the scene. The control of annotation granularity also improves the quality as it may reduce some annotations like \textit{woman-has-hair} introduced by the noisy and overlapping bounding boxes.

Along with the PSG dataset, they propose two baselines, including PSGTR and PSGFormer. PSGTR~\cite{psg} used a triplet query to model the relations in the scene graph as \texttt{{{Subject}, \texttt{Relation}, \texttt{Object}}} pairs, while PSGFormer~\cite{psg} applied both object query and relation query to model the nodes and edges in the scene graph separately, then applied a relation-based fetching to find the most relevant object queries through some interaction modules to build the scene graph. Nonetheless, without explicit modeling of objects, the triplet pair approaches require heavy hand-designed post-processing modules to merge all triplets into a single graph, which may fail to keep the consistent entity-relation structure. As for relation-based fetching, such an approach is not effective and straightforward.

\vspace{2mm}
\noindent
\textbf{Panoptic Segmentation.} This task unifies the semantic segmentation and instance segmentation into one framework with a single metric named Panoptic Quality (PQ)~\cite{pq}. 
Lots of work has been proposed to solve this task using various approaches. However, most works~\cite{pan-sepa-1,pan-fcn,pan-fpn,upsnet,autopan} separate thing and stuff prediction as individual tasks. Recently, several approaches~\cite{mask2former,knet,maskformer,video-knet, yuan2022polyphonicformer,li2023panopticpartformer++,li2023tube,han2023reference,yang2021collaborative,li2023catr,xu2023video} unify both thing and stuff prediction as a mask-based set prediction problem. Our method is based on the unified model. However, as shown in \Cref{tab:pair_recall}, better segmentation quality does not mean a better panoptic scene graph result. We pay more attention to the panoptic scene graph generation, with the main focus on pair-wised relation detection.

\vspace{2mm}
\noindent
\textbf{Detection Transformer.} Starting from DETR~\cite{detr}, object query-based detectors~\cite{sparse_rcnn, deformable_detr} are designed using object queries to encode each object and model object detection as a set prediction problem. Several approaches generalize the idea of using object queries for other domains, such as segmentation~\cite{max-deeplab, maskformer, mask2former}, tracking~\cite{tracking1, tracking2, tracking3, tracking4, video-knet, li2023tube}, and scene graph generation~\cite{sgtr, psg, pst}. In particular, SS-RCNN~\cite{structured-rcnn} uses triplet queries to directly output sparse relation detections. However, the relationship between objects and subjects is not explicitly learned or well explored. Moreover, it can not generalize into PSG directly since the limited mask resolution results by RoI align~\cite{faster_rcnn}.

\section{Methods}
\label{sec:method}
In this section, we will first formulate the problem setting of PSG in~\Cref{sec:problem_setting}. Then, we will present our findings from three different aspects in~\Cref{sec:motivation}: enough capability of current segmenters, the importance of pair recall, and the sparsity of pair-wise relations. Following this, we will illustrate the details of our Pair-Net architecture in~\Cref{sec:pair-net}, including the Panoptic Segmentation Network, Pair Proposal Network, and Relation Fusion module. Finally, we will provide the training and inference procedures in~\Cref{sec:long-tail}.

\begin{table}
    \centering
    \caption{\textbf{IoU and Recall$_{\mathbf{0.5}}$ of COCO-pretrained detectors on PSG.} IoU and Recall$_{0.5}$ are averaged at the triplet level of the scene graph. It shows the excellence of object-level recall.}
    \label{tab:detector_iou}
    \scalebox{1.0}{
        \begin{tabular}{c|cc|cc}
            \toprule[0.15em]
            Model                          & sub-IoU & obj-IoU & sub-R$_{0.5}$ & obj-R$_{0.5}$ \\
            \midrule[0.15em]
            DETR \cite{detr}               & 0.74    & 0.73    & 0.87          & 0.84          \\
            Mask2Former \cite{mask2former} & 0.79    & 0.78    & 0.91          & 0.90          \\
            \bottomrule
        \end{tabular}
    }
\end{table}

\begin{table}
    \centering
    \caption{\textbf{Pair recall@20, triplet recall@20 and PQ of different models on PSG.} PSGFormer$^{+}$ denotes that the detector of PSGFormer is changed from DETR to Mask2Former. The table shows that different models have a similar ability in panoptic segmentation (PQ), but Pair Recall is strongly correlated to Triplet Recall.}
    \label{tab:pair_recall}
    \scalebox{1.0}{
        \begin{tabular}{c|ccc}
            \toprule[0.15em]
            Model                                   & Pair R@20     & R@20          & PQ            \\
            \midrule[0.15em]
            MOTIFS \cite{motifs}                    & 36.7          & 20.0          & 40.4          \\
            VCTree \cite{vctree}                    & 37.2          & 20.6          & 40.4          \\
            GPS-Net \cite{gpsnet}                   & 34.3          & 17.8          & 40.4          \\
            PSGFormer \cite{psg}                    & 26.6          & 18.0          & 36.8          \\
            PSGFormer$^{+}$ \cite{psg, mask2former} & 28.6          & 18.9          & 43.8          \\
            \cmidrule(){0-3}
            \textbf{Pair-Net (Ours)}                & \textbf{52.7} & \textbf{29.6} & \textbf{40.2} \\
            \bottomrule
        \end{tabular}
    }
\end{table}

\noindent
\subsection{Problem Setting}
\label{sec:problem_setting}
Following the previous literatures~\cite{sgg1, sgg_survey1,sgg_survey2,sgg_survey3},
a scene graph $\mathbf{S}$ is defined as a set of \texttt{\{Subject}, \texttt{Relation}, \texttt{Object\}} triplets. In the task of Scene Graph Generation (SGG), the distribution of a scene graph can be summarized in the following two steps:

\begin{equation}
    \label{eq:sgg}
    p(\mathbf{S} | \mathbf{I}) = p(\mathbf{B}, \mathbf{A} | \mathbf{I}) p(\mathbf{R} |\mathbf{B}, \mathbf{A}, \mathbf{I})
\end{equation}

Here, $\mathbf{I} \in \mathbb{R}^{H \times W \times 3}$ is the input image, and $\mathbf{B}=\left\{b_{1}, \ldots, b_{n}\right\}$ and $\mathbf{A}=\left\{a_{1}, \ldots, a_{n}\right\}$ represent the bounding box coordinates and class labels of $n$ objects in the image, respectively. $\mathbf{R}=\left\{r_{1}, \ldots, r_{m}\right\}$ represents the relations between objects given $\mathbf{B}$ and $\mathbf{A}$. Here, $a\in\mathbb{Z}^x$ and $r\in\mathbb{Z}^y$ refer to one of the pre-defined $x$ object and $y$ relation classes. The values of $n$ and $m$ are arbitrary numbers depending on the context of the image.

Instead of localizing only (foreground) objects using the annotation at the level of bounding box coordinates, Panoptic Scene Graph Generation (PSG) grounds each foreground object (thing) with the more fine-grained panoptic segmentation mask annotation. It incorporates information from background objects (stuff) into the consideration of a scene graph. Formally, the model is trained to learn the following distribution:

\begin{equation}
    \label{eq:psg}
    p(\mathbf{S} | \mathbf{I}) = p(\mathbf{M}, \mathbf{A} | \mathbf{I}) p(\mathbf{R} |\mathbf{M}, \mathbf{A}, \mathbf{I})
\end{equation}

In this case, $\mathbf{M}=\left\{m_{1}, \ldots, m_{n}\right\}$ and $m_{i} \in{0,1}^{H \times W}$ represent the panoptic segmentation masks corresponding to both foreground and background objects in the image.

\subsection{Motivation}
\label{sec:motivation}

\noindent
\textbf{Query-based Segmenter is Good Enough for PSG.}
To learn the pair-wised relation between different entities in PSG, we first study the question, \textit{`whether the query-based segmenter can encode semantic information of corresponding subject or object using object queries?'} In this way, scene graph generation could be simplified into learning the pair-wised relationship between subjects and objects and classifying object queries to subject and object, respectively. We use COCO-pre-trained models to test both mean mask IoU and Recall of each subject and object depicted in the scene graph. The results, presented in \Cref{tab:detector_iou}, show that both DETR~\cite{detr} and Mask2Former~\cite{mask2former} are effective in recalling. Recall$_{0.5}$ indicates a prediction is correct if the IoU between it and the ground truth is larger or equal to $0.5$, which is the common practice used in the segmentation community. Given their high Recall$_{0.5}$ and IoU, we argue that the quality of panoptic segmentation is already good enough to support the following pair then relation generation, and the performance of PSG models is not bounded by object segmenters. Additionally, this also suggests that object queries, which are used for mask and class prediction, can be utilized as a good predicate to learn the pairwise relationships between entities in PSG directly.

\vspace{2mm}
\noindent
\textbf{Better Pair Recall, Better Triplet Recall in PSG.} In \Cref{tab:pair_recall}, we test different PSG methods in cases of their recall (main metric of PSG) and pairwise recall. The pair recall is calculated by jointly considering object and subject predictions and omitting the relation classification correctness, which shares the similar thought of Region Proposal Network (RPN)~\cite{faster_rcnn} to recall all possible foreground objects. We find pair recall is more important than segmentation quality, while different methods with similar PQ have various recalls. This motivates us to design a model to directly enhance the recall of PSG in pairs rather than improving segmentation quality, which is already good enough for SGG tasks.

\begin{figure*}[!ht]
    \centering
    \includegraphics[width=1.0\linewidth]{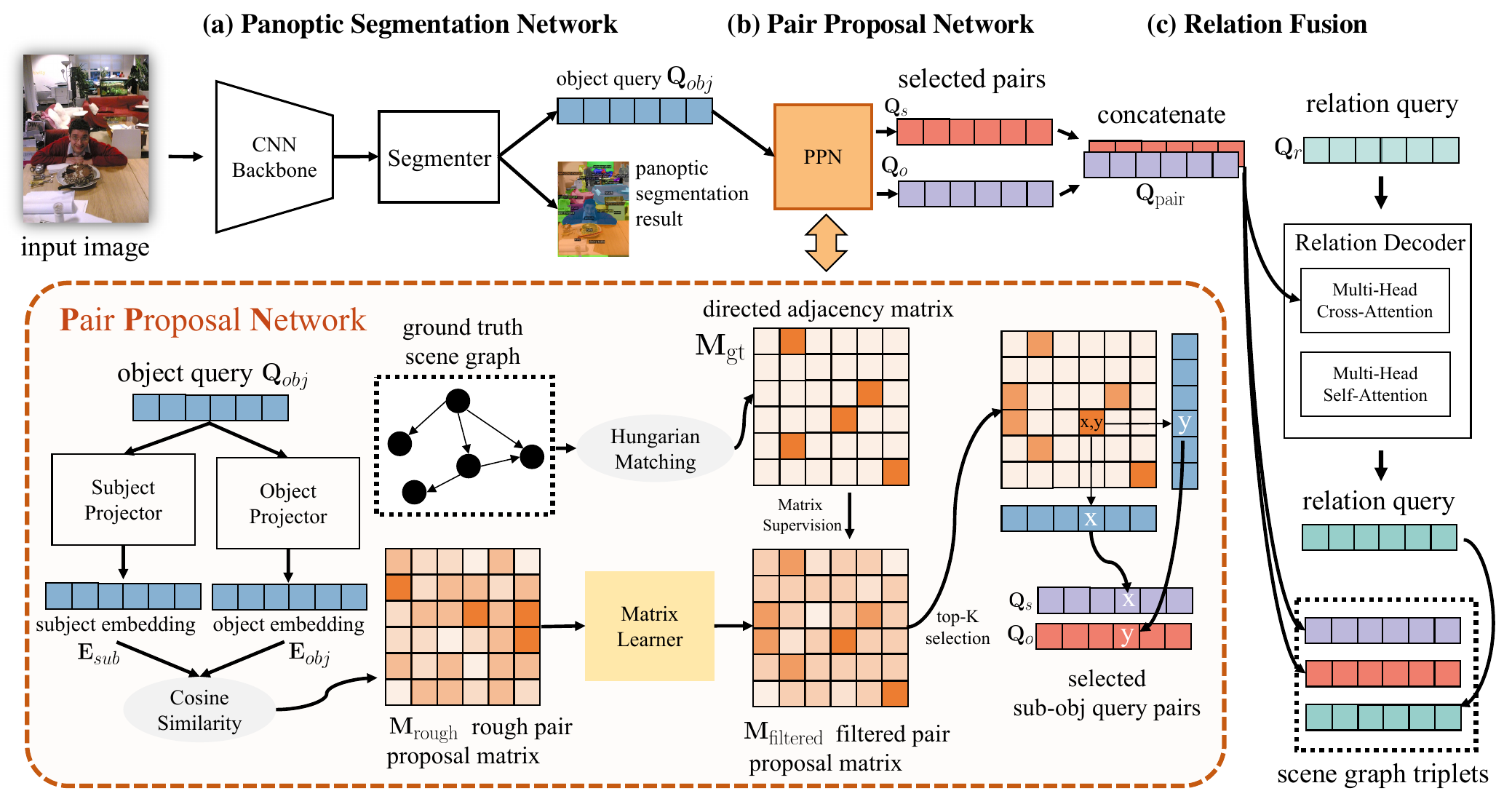}
    \caption{\textbf{An illustration of our proposed Pair-Net.} It mainly contains three parts: (a) Panoptic Segmentation Network uses a query-based object segmenter to generate panoptic segmentation and object queries. (b) Pair Proposal Network generates subject-object pairs from object queries, with Matrix Learner to ensure the sparsity property. (c) The relation Fusion module models the interaction between pair-wised queries and relation queries and predicts final relation labels.}
    \label{fig:overall}
\end{figure*}

\subsection{Pair-Net Architecture}
\label{sec:pair-net}

As shown in \Cref{fig:overall}, our Pair-Net mainly contains three parts. Firstly, we use the Mask2Former baseline to extract the object queries. Then, we use the proposed Pair Proposal Network to recall all confident subject-object pairs and select the best pairs. Finally, we use the Relation Fusion module to decode the final relation prediction between subjects and objects.

\subsubsection{Panoptic Segmentation Network}
\label{sec:seg_network}

We adopt strong Mask2Former~\cite{mask2former} as our segmenter in the panoptic segmentation network. Mask2Former contains a transformer encoder-decoder architecture with a set of object queries, where the object queries interact with encoder features via masked cross-attention. Unlike RelPN~\cite{relpn} separately generates proposals for subjects, objects, and relations with bounding boxes using 3 branches, the segmenter jointly produces panoptic segmentation of the subjects and objects, without consideration of relation. Given an image $\mathbf{I}$, during the inference, the Mask2Former directly outputs a set of object queries $\mathbf{Q_\text{obj}}=q_{\{i\}}, i=1 \ldots N$, where each object query $q_{i}$ represent one entity. We denote it as $\mathbf{Q}_\text{obj} \in \mathbb{R}^{N_\text{obj}\times d}$, where $N_{obj}$ is the number of object queries and $d$ is the embedding dimensions. Each object query is matched to ground truth masks during training via masked-based bipartite matching. The loss function is $\mathcal{L}_\text{mask} = \lambda_{cls}\mathcal{L}_{cls} + \lambda_{ce}\mathcal{L}_{ce} + \lambda_{dice}\mathcal{L}_{dice}$, where $\mathcal{L}_{cls}$ is Cross Entropy (CE) loss for mask classification, and $\mathcal{L}_{ce}$ and $\mathcal{L}_{dice}$ are CE loss and Dice loss~\cite{dice_loss} for segmentation, respectively.

\subsubsection{Pair Proposal Network}
\label{sec:ppn}

Our Pair Proposal Network (PPN) focuses on predicting the relative importance of subject/object queries and then selects top-k subject-object pairs according to the index of the top-k value in the pair proposal matrix.

As shown in~\Cref{fig:overall}, our PPN consists of a subject projector, an object projector, and a matrix learner. The projector layer is an MLP that will generate subject and object embedding $\mathbf{E}_{\text{sub}},\ \mathbf{E}_{\text{obj}} \in \mathbb{R}^{N_\text{obj}\times d}$ respectively from input $\mathbf{Q}_\text{obj}$. After that, cosine similarity between $\mathbf{E}_{\text{sub}}$ and $\mathbf{E}_{\text{obj}}$ is calculated, i.e., a rough sketch of Pair Proposal Matrix $\mathbf{M}_{\text{rough}} \in \mathbb{R}^{N_\text{obj} \times N_\text{obj}}$. Finally, a Matrix Learner is applied to further filter the rough sketch of Pair Proposal Matrix $\mathbf{M}_{\text{rough}}$, generating a more precise prediction of importance in the Pair Proposal Matrix. To avoid ambiguity, such interaction is the pairing step between objects, following our pair-then-relation generation order. It is not relevant to the design of RelPN~\cite{relpn}, which directly calculates the visual and spatial compatibility among subjects, objects, and relations.

\vspace{2mm}
\noindent\textbf{Matrix Learner.} Taking the motivation from~\Cref{sec:motivation}, a small network, namely matrix learner, is designed to do further filtration and learn feasible sparse pairs. In particular, we find using simple CNN architecture could achieve better results. Rather than using transformer architecture like ViT~\cite{vit}, we argue that a CNN architecture can preserve the local details while filtering the redundant noise as an efficient semantic filter. Its output is a filtered pair proposal matrix $\mathbf{M}_{\text{filtered}}$, which contains the sparser connectivity representation compared to its input matrix $\mathbf{M}_{\text{rough}}$. Finally, a top-k selection is conducted on the $\mathbf{M}_{\text{filtered}}$. The result top-k indexes are used to select the corresponding subject and object queries from $\mathbf{Q}_\text{obj}$, which are denoted as $\mathbf{Q}_{s}$ and $\mathbf{Q}_{o}$ respectively. This selection process means partial of the $\mathbf{Q}_\text{obj}$ will be processed into the next module but does not mean the gradient flow will be interrupted for the end-to-end computation. To supervise the filtration process of the matrix learner, we introduce additional information about pair-wised relations from the ground truth, which is defined as $\mathbf{M}_\text{gt} \in \mathbb{R}^{N_\text{obj} \times N_\text{obj}}$. Using Hungarian matching, we can assign each subject and object in the ground truth scene graph to a specific object query $q_k$ based on their segmentation losses. After all ground-truth subject-object pairs have been assigned, multiple positions of $\mathbf{M}_\text{gt}$ will be assigned to $1$, and the remaining positions will be assigned to $0$.

We use such a matrix to supervise the Matrix Learner with a Binary Cross Entropy loss (BCE). Due to the sparsity of $\mathbf{M}_\text{gt}$, we enhance the BCE loss with a positive weight adjustment to ensure stable training. On average, only 5.6 relations are expected in one image in the PSG dataset~\cite{psg}. 
Therefore, in the ground truth matrix $\mathbf{M}_\text{gt} \in \mathbb{R}^{N_\text{obj} \times N_\text{obj}}$ ($N_\text{obj} = 100$ in our case), it is expected that only $5\sim6$ positions will have a value of $1$, while other positions have a value of $0$. 
This supervision could be considered as a very unbalanced binary classification problem. 
To encourage the Matrix Learner not to output an all-zero $\mathbf{M}_\text{filtered}$, we dramatically increase the weight $p$ of the loss of the positive position, which introduces us to the positive weight adjustment BCE loss.
This loss forces the network to produce sparse relationships for both subject and object pairs, and we derive our proposal pair loss as:

\begin{equation}
    \mathcal{L}_\mathbf{ppn} = -\left[p\cdot y \cdot \log{\sigma(x)} + (1-y)\cdot\log(1-\sigma(x))\right]
\end{equation}

where $p$ = $\frac{N^2_\text{obj}}{\sum_{i,j}\mathbf{M}_\text{gt}(i,j)}$, \text{i.e., the reciprocal of $\mathbf{M}_\text{gt}$'s sparsity.}

\subsubsection{Relation Fusion}
\label{sec:rel-fusion}
After selecting the top-k queries, we adopt another Transformer decoder to predict their relations. As shown in~\Cref{fig:overall} (c), we term it as Relation Fusion. In this module, we have a relation decoder consisting of transformer decoders in the style of~\cite{detr}. After selecting sub-query $\mathbf{Q}_{s}$ and obj-query $\mathbf{Q}_{o}$ from the object query $\mathbf{Q}_\text{obj}$ via PPN, they are concatenated together along the length dimension to construct a pair query $\mathbf{Q}_{\text{pair}} \in \mathbb{R}^{2N_\text{rel}\times d}$, which are projected as the key and value of cross attention in the relation decoder. We randomly initialize a trainable relation query $\mathbf{Q}_r \in \mathbb{R}^{N_\text{rel}\times d}$ as the query input. $N_\text{rel}$ denotes the number of relation queries and $d$ denotes the embedding size of the decoder. $N_\text{rel}$ shall be the same as $k$ as a valid triplet has to be formed by one object, one subject, and one relation. And, it is worth noticing that this value is decoupled with the number of the output object query of the segmenter (i.e., $100$ in Mask2Former).

The cross attention mechanism \cite{attention} between $\mathbf{Q}_r$ and $\mathbf{Q}_{\text{pair}}$ yields an equivalent matching effect through the dot product in the attention formulation. Such that, the $i^\text{th}$ relation query mainly pays its attention to the $i^\text{th}$ of the $\mathbf{Q}_{\text{pair}}$ while still gaining some information from the other pairs. Since the relation query is in the same order as the pair query, \textit{no further post-processing or matching} between pairs and relations is needed in this stage. The output of the relation fusion module is predicted relation classes. We illustrate the loss of the relation fusion module in the following section.

\subsection{Training and Inference}
\label{sec:long-tail}

\begin{figure}[t]
    \centering
    \includegraphics[width=0.99\linewidth]{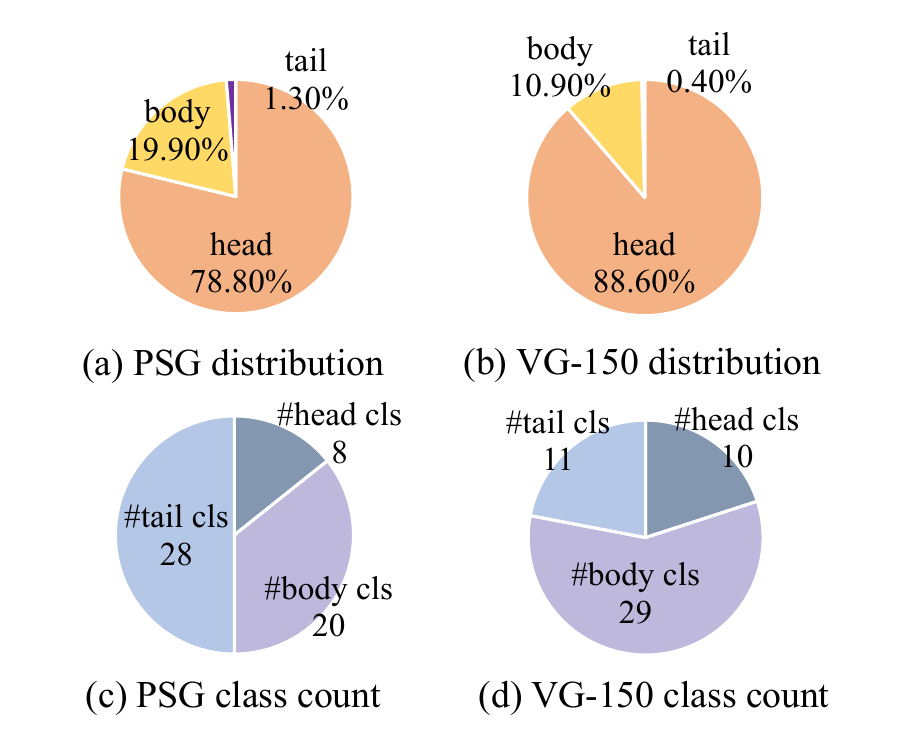}
    \caption{\textbf{Relation classes distribution of PSG and VG-150.} Following~\cite{bi-graph}, we summarize the proportion and number of different classes in the form of \textit{head, body, tail} of PSG in (a) and (c). We provide results of VG-150 in (b) and (d) for reference. The figure shows the long-tail effect on the distribution of relation classes.}
    \label{fig:rel-cls}
\end{figure}

\noindent
\textbf{Long-tailed Distribution on Relation Classes.} As SGG tasks, we have observed a long-tailed distribution of relation classes, which can heavily affect the performance of mean AR. Following the definition in previous literature~\cite{bi-graph}, we split all relation classes into three disjoint groups according to their counts in the training split: head ($>$10k), body (500 $\sim$ 10k), and tail ($<$500). From~\Cref{fig:rel-cls}, we notice that half of the relation classes (tail classes) only account for about $1\%$ and eight classes (head classes) account for about $80\%$ in the training set in PSG. This clearly indicates the characteristic of a long-tailed distribution in the relation classes. 
There are several methods to handle long-tailed distributions. At the dataset level, resampling of the original dataset with logit adjustment of relation classes can be applied, generating an augmented dataset with a more balanced distribution of relations. At the loss level of the relation fusion module, Focal Loss~\cite{focalloss} modifies the standard cross entropy loss and applies weighted discrimination on the well-classified classes, which forces the model to focus on wrongly classified classes. Furthermore, Seesaw loss~\cite{seesawloss} dynamically re-balances gradients of positive and negative samples, which is adopted for relation classification in our framework by default. All these different methods will affect the performance of relation loss $\mathcal{L}_{r}$. Among these three choices, we select Seesaw loss in our default training setting as we find it yields the best performance in our experiments.

\begin{figure}[t]
    \centering
    \includegraphics[width=0.8\linewidth]{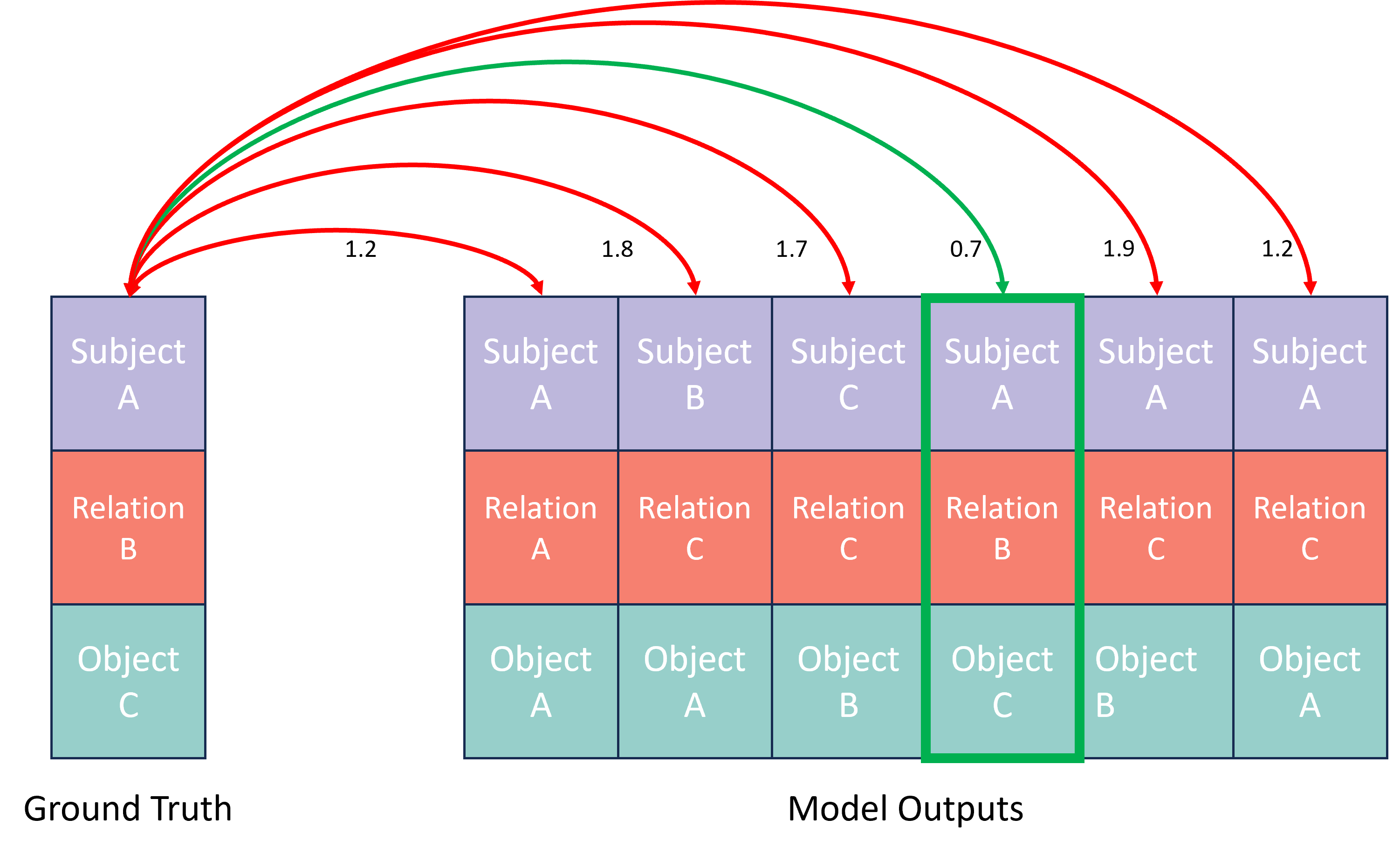}
    \caption{\textbf{Illustration of Hungarian matching process before the losses calculation for one ground truth example.} We define the matching cost as the sum of subject, object, and relation classification loss between the model output and the ground truth. The model outputs are the probabilities of all possible classes, and we illustrate the label of the highest value of the distribution for simplicity.}
    \label{fig:hungarain_matching}
\end{figure}

\vspace{2mm}
\noindent
\textbf{Training Loss.} Our Pair-Net can be trained as a one-stage SGG model. The entire loss contains three classification losses for the subject, object, and relation, one binary classification loss for PPN, and the origin mask loss of Mask2Former.
For the calculation of the three classification losses for the subject, object, and relation, we follow the common practice of optimal bipartite matching proposed in DETR~\cite{detr} by matching the $K$ outputs given by the model and the $N$ ground-truth triplets ($N<K$). We define the Hungarian matching cost as the sum of classification losses and illustrate a sample in \Cref{fig:hungarain_matching}.
Overall, the loss of our framework is defined as:
%
\begin{equation}
    \begin{aligned}
        \mathcal{L} & =\lambda_\text{subject classification}\,  \mathcal{L}_\text{subject classification}
        \\ &+ \lambda_\text{object classification}\, \mathcal{L}_\text{object classification}
        \\ &+ \lambda_\text{relation classification}\,  \mathcal{L}_\text{relation classification}
        \\ &+ \lambda_\text{pair proposal network}\,  \mathcal{L}_\text{ppn}
        \\ &+
        \lambda_\text{original Mask2Former Loss}\, \mathcal{L}_\text{original Mask2former Loss}
    \end{aligned}
\end{equation}
where we set $\lambda_\text{o}=\lambda_\text{s} = 4, \lambda_\text{r}= 2, \lambda_\text{ppn}=5, \lambda_\text{original}=1$.\\
\noindent
\textbf{Inference.} The model takes an image as input. Firstly, the segmenter produces object queries $\mathbf{Q}_\text{obj}$, the object classification result, and the mask segmentation result. Then, the PPN selects obj-query $\mathbf{Q}_o$ and sub-query $\mathbf{Q}_s$ based on the top-k index of the filtered pair proposal matrix $\mathbf{M}_\text{filtered}$. After concatenation, the selected $\mathbf{Q}_o$ and $\mathbf{Q}_s$ form as pair query $\mathbf{Q}_\text{pair}$ as the key and value function of the relation decoder. Finally, the relation decoder produces a relation query, which is fed into a single linear layer for relation classification. The result triplets are given by concatenating subject, object, and relation classification results.

\begin{table}[!t]
    \centering
    \caption{\textbf{Different setting comparison with previous SGG and PSG methods.}}
    \scalebox{0.77}{{\begin{tabular}{c | c c | c  c  c c }
                    \toprule[0.15em]
                    Method                                         & SGG        & PSG        & End-to-end & Sparsity   & Pair Loss  & Background \\
                    \midrule[0.15em]
                    \rowcolor{gray!15} RelPN~\cite{relpn}          & \checkmark &            &            &            &            &            \\
                    \rowcolor{gray!15} Graph-RCNN~\cite{graphrcnn} & \checkmark &            &            & \checkmark &            &            \\
                    \rowcolor{gray!15} SGTR~\cite{sgtr}            & \checkmark &            & \checkmark & \checkmark &            &            \\
                    \hline
                    \rowcolor{orange!15} PSGTR~\cite{psg}          &            & \checkmark & \checkmark &            &            & \checkmark \\
                    \rowcolor{orange!15} PSGFormer~\cite{psg}      &            & \checkmark & \checkmark &            &            & \checkmark \\
                    \hline
                    \rowcolor{red!15} \textbf{Pair-Net (Ours)}     & \checkmark & \checkmark &            & \checkmark & \checkmark & \checkmark \\
                    \bottomrule
                \end{tabular}}}
    \label{tab:more_detailed_comparison}
\end{table}

\begin{table*}
    \centering
    \caption{\textbf{Results on PSG validation dataset.} Pair-Net outperforms previous methods by a large margin on all metrics. Our model outperforms prior state-of-the-art models by an absolute 10.2\% in mR@20 and 11.6\% in R@20. The $^{+}$ mark indicates our re-implementation of previous methods with Mask2Former as the detector. All results are given by training 12 epochs except PSGTR, which needs 60 epochs to converge. Previous SGG methods are adapted to the PSG task by using their scene graph generation modules to process object features from classic panoptic FPN.}
    \label{tab:main1}
    \resizebox{\textwidth}{!}{
        \begin{tabular}{c|c|c|@{\hskip 15pt}c@{\hskip 15pt}c@{\hskip 15pt}c@{\hskip 15pt}|@{\hskip 15pt}c@{\hskip 15pt}c@{\hskip 15pt}c@{\hskip 15pt}}
            \toprule[0.15em]
            BackBone                               & Detector                                       & Model                              & mR@20 & mR@50 & mR@100 & R@20 & R@50 & R@100 \\
            \midrule[0.15em]
            \multirow{8}*{ResNet-50~\cite{resnet}} & \multirow{4}*{Faster R-CNN~\cite{faster_rcnn}} & IMP \cite{imp}                     & 6.5   & 7.1   & 7.2    & 16.5 & 18.2 & 18.6  \\
                                                   &                                                & MOTIFS \cite{motifs}               & 9.1   & 9.6   & 9.7    & 20.0 & 21.7 & 22.0  \\
                                                   &                                                & VCTree \cite{vctree}               & 9.7   & 10.2  & 10.2   & 20.6 & 22.1 & 22.5  \\
                                                   &                                                & GPS-Net \cite{gpsnet}              & 7.0   & 7.5   & 7.7    & 17.8 & 19.6 & 20.1  \\
            \cmidrule(){2-9}
                                                   & \multirow{2}*{DETR~\cite{detr}}                & PSGFormer \cite{psg}               & 14.5  & 17.4  & 18.7   & 18.0 & 19.6 & 20.1  \\
                                                   &                                                & PSGTR \cite{psg}         & 16.6  & 20.8  & 22.1   & 28.4 & 34.4 & 36.3  \\

            \cmidrule(){2-9}
                                                   & \multirow{6}*{Mask2Former~\cite{mask2former}}  & PSGFormer$^{+}$~\cite{psg}         & 16.6  & 19.4  & 20.3   & 18.9 & 21.5 & 22.4  \\
                                                   &                                                & PSGTR$^{+}$~\cite{psg}   & 20.9  & 27.4  & 28.4   & 32.6 & 38.0 & 38.9  \\
                                                   &                                                & HiLo~\cite{zhou2023hilo} & 23.7  & 30.3  & 33.1   & 34.1 & 40.7 & 43.0  \\
            \cmidrule(){3-9}
                                                   &                                                & \textbf{Pair-Net (Ours)}           & 24.7  & 28.5  & 30.6   & 29.6 & 35.6 & 39.6  \\
            \cmidrule{1-1}
            Swin-B~\cite{swin}                     &                                                & \textbf{Pair-Net$^{\dag}$ (Ours)}  & 25.4  & 28.2  & 29.7   & 33.3 & 39.3 & 42.4  \\
            \bottomrule
        \end{tabular}}
\end{table*}

\subsection{Relation With Previous SGG Methods}

We provide a more detailed description and comparison of Pair-Net in \Cref{tab:more_detailed_comparison} with several feature comparisons. Compared with previous works, our proposed Pair-Net supports both SGG and PSG. Moreover, we adopt pair loss to directly optimize the relation between different object and subject pairs, which is different from all previous works. As a result, our method leads to sparsity in relation pairs within the query-based framework. Compared with ReIPN~\cite{relpn} and Graph-RCNN~\cite{graphrcnn}, our method also considers the background context.
\section{Experiment}
\label{sec:exp_section}

\noindent
\textbf{Panoptic Scene Graph (PSG) Benchmark~\cite{psg}.} Filtered from COCO~\cite{coco} and VG datasets \cite{vg}, the PSG dataset contains 133 object classes, including things, stuff, and 56 relation classes. This dataset has 46k training images and 2k testing images with panoptic segmentation and scene graph annotation. We follow the same data processing pipelines from \cite{psg}.

\vspace{2mm}
\noindent
\textbf{Evaluation Metrics.} There are multiple evaluation matrices to evaluate the performance of a scene graph model. 
The Predicate Classification (PredCls) and Scene Graph Classification (SGCls)~\cite{kern} take truth object classification and localization labels as input and output their pairwise relation. 
Scene Graph Generation (SGGen) is the hardest one, which asks the model to predict correct triplets and object segmentation simultaneously. 
The IoU threshold of a correct mask is set to 0.5, and a correct matching means all elements in the triplet \texttt{\{Subject}, \texttt{Relation}, \texttt{Object\}} are classified correctly. We report recall@K (R@K) and mean recall@K (mR@K) for K $=20, 50, 100$ following the definition from~\cite{vctree}. 
R@K indicates that calculating the recall given the top-k-confident triplet classification results from the model. mR@K means the average of the recall of each relation class following the above procedure. \footnote{For more detailed illustration, one might check the \href{https://github.com/KaihuaTang/Scene-Graph-Benchmark.pytorch/blob/master/METRICS.md}{Github repo: Scene-Graph-Benchmark.pytorch}}

\vspace{2mm}
\noindent
\textbf{Training Configuration.} For the panoptic segmentation task in PSG, we use COCO~\cite{coco} pre-trained Mask2Former\footnote{Since PSG is a subset of COCO, it is equivalent to pre-train the segmenter on the PSG.} as the object segmenter. Our framework is optimized by AdamW \cite{adamw} with an initial learning rate of $10^{-4}$, a weight decay of $10^{-4}$, and a total batch size of $8$. We train Pair-Net for a total of 12 epochs and reduce the learning rate by a factor of 0.1 at epoch 5 and 10. We set all the positional encoding of the query, key, and value in the Relation Fusion module learnable. We conducted 5 experiments with different seeds and the results are stable given $\pm 0.2$ in terms of mR@20.

\vspace{2mm}
\noindent
\textbf{General Framework Hyperparameters.} We set the number of object queries to $N_{obj} = 100$, size of embedding dimensions, $d = 256$ inheriting the design of Mask2Former~\cite{mask2former}. The subject projector and object projector are both MLPs with three fully connected layers, with embedding dimension $d=256$ and ReLU as the activation function. For the Matrix Learner, we use a 3-layer CNN with 64 inner channels and a 7 by 7 kernel size. The Relation Fusion module consists of a 6-layer DETR-style~\cite{detr} transformer self-attention and cross-attention decoder with $d=256$. 

\subsection{Main Results}

\noindent
\textbf{Comparison with Baselines.} The previous two-stage models, all of them, choose Faster R-CNN \cite{faster_rcnn} as Detectors. For a fair comparison, we create a stronger baseline method based on PSGFormer and PSGTR ~\cite{psg} using Mask2Former~\cite{mask2former} as Detector noted as PSGFormer$^{+}$. As shown in~\Cref{sec:motivation}, the segmenter can detect and segment each subject and object, including \textit{thing} and \textit{stuff}. In~\Cref{tab:main1}, we apply Recall@20/50/100 and Mean Recall@20/50/100 as our benchmarks. All models use ResNet-50 for a fair comparison. As shown in~\Cref{tab:main1}, our method Pair-Net achieves 29.6\% in R@20 and outperforms the baseline by a 10.7\% large margin. Additionally, for R@50 and R@100, the margin is even larger, which proves that our methods can better utilize all 100 proposals and provide reasonable and not self-repeated predictions. Considering mean Recall, our method also outperforms the baseline with absolute $10.2 \sim 11.9$ gain for $K=20,50,100$.

\vspace{2mm}
\noindent
\textbf{Pair Recall Improvement.} In~\Cref{tab:pair_recall}, our Pair-Net also gains significant improvement on the Pair Recall@20 compared with existing methods. A large 24.1\% margin on Pair Recall@20 is obtained compared with the baseline method, PSGFormer$^{+}$. This observation strengthens our assumption that Pair Recall is highly correlated to Recall, and it is a current bottleneck for the PSG model performance.

\vspace{2mm}
\noindent
\textbf{Categorical Recall@K on PSG.} In PSG, which benefited from the panoptic segmentation setting, the relation is constructed not only from thing to thing (TT) class but also from stuff to stuff (SS), stuff to thing (ST), and thing to stuff (TS). To this end, we introduce four new different metrics to evaluate the performance of the model further: \textbf{TT-Recall@K}, \textbf{SS-Recall@K}, \textbf{ST-Recall@K}, and \textbf{TS-Recall@K}. They calculate the recall on the four categories independently. From~\Cref{tab:c_recall}, our Pair-Net mainly improves the recall of all the cases. Our findings suggest that we should pay more attention to \textit{pair recall for PSG} rather than improving segmentation quality.

\vspace{2mm}
\noindent
\textbf{Stronger Backbone for Pair-Net}. We train a larger Pair-Net with Swin-Base \cite{swin} as the backbone for future research. A larger backbone could further improve the performance of Pair-Net with absolute $\sim 1\%$ in mean recall and $3\sim 4\%$ in Recall. This indicates the scalability of Pair-Net.

\begin{table}
    \centering
    \caption{\textbf{Four categorical Recall (R)@20 in PSG.} We introduce four categorical recalls and report the performance in terms of Recall@20.}
    \label{tab:c_recall}
    \scalebox{0.9}{
        \begin{tabular}{c|cccc}
            \toprule[0.15em]
            Model                                   & TT-R@20 & TS-R@20 & ST-R@20 & SS-R@20 \\
            \midrule[0.15em]
            PSGFormer~\cite{psg}                    & 17.2    & 21.7    & 14.9    & 14.7    \\
            PSGFormer$^{+}$~\cite{psg, mask2former} & 19.5    & 21.5    & 9.5     & 18.5    \\
            Pair-Net (ours)                         & 25.7    & 31.5    & 24.2    & 34.2    \\
            \bottomrule
        \end{tabular}}
\end{table}

\begin{table*}[!ht]
    \centering
    \caption{\textbf{Pair-Net ablation experiments on PSG.} We report mean Recall and Recall with K=20, 50. Our settings are marked in \colorbox{lightgray!50}{gray}.}
    \subfloat[Necessity of Each Component of the PPN.]{
        \label{tab:ablation_a}
        \hfill
        \begin{tabularx}{0.58\textwidth}{ccc|cc}
            \toprule[0.15em]
            Linear Embed & Matrix Learner & BCE supervision & mR/R@20                             & mR/R@50                             \\
            \toprule[0.15em]
            \checkmark   & \checkmark     & $\times$        & 0.5 / 0.4                           & 1.3 / 1.2                           \\
            \checkmark   & $\times$       & \checkmark      & 14.8 / 20.5                         & 21.0 / 29.8                         \\
            $\times$     & \checkmark     & \checkmark      & 14.6 / 22.1                         & 17.8 / 27.9                         \\
            \checkmark   & \checkmark     & \checkmark      & \cellcolor{lightgray!50}24.7 / 29.6 & \cellcolor{lightgray!50}28.5 / 35.6 \\
            \bottomrule
        \end{tabularx}
    }\hfill
    \subfloat[Different Architectures for Matrix Learner.]{
        \label{tab:ablation_b}
        \begin{tabularx}{0.38\textwidth}{c|c c c}
            \midrule[0.15em]
            Architecture & \# Para                      & mR/R@20                             & mR/R@50                             \\
            \toprule[0.15em]
            MLP          & 0.2M                         & 13.0 / 18.8                         & 19.4 / 26.1                         \\
            G-Transformer & 1.4M & 17.3 / 26.1 & 23.2 / 32.6\\
            W-Transformer & 1.3M & 15.5 / 21.4 & 19.4 / 26.7 \\
            CNN-tiny     & \cellcolor{lightgray!50}0.2M & \cellcolor{lightgray!50}24.7 / 29.6 & \cellcolor{lightgray!50}28.5 / 35.6 \\
            CNN-base     & 30M                          & 23.3 / 33.3                         & 28.2 / 39.3                         \\
            \bottomrule
        \end{tabularx}
    } \hfill
    \subfloat[Different loss functions for relation classification.]{
        \label{tab:ablation_c}
        \begin{tabularx}{0.36\textwidth}{c|c c}
            \toprule[0.15em]
            Method                        & mR/R@20                             & mR/R@50                             \\
            \midrule[0.15em]
            Cross Entropy Loss            & 15.4 / 29.0                         & 17.0 / 34.0                         \\
            Weighted Resampling           & 12.6 / 21.4                         & 17.6 / 29.3                         \\
            Focal Loss \cite{focalloss}   & 19.8 / 28.3                         & 22.1 / 33.6                         \\
            Seesaw Loss \cite{seesawloss} & \cellcolor{lightgray!50}24.7 / 29.6 & \cellcolor{lightgray!50}28.5 / 35.6 \\
            \bottomrule
        \end{tabularx}
    }\hfill
    \subfloat[Different KV inputs for relation fusion decoder.]{
        \label{tab:ablation_d}
        \begin{tabularx}{0.305\textwidth}{c|c c} \toprule[0.15em]
            Input          & mR/R@20                             & mR/R@50                             \\
            \midrule[0.15em]
            random init    & 7.3 / 0.7                           & 9.8 / 1.0                           \\
            image features & 1.3 / 2.3                           & 2.6 / 4.1                           \\
            random pairs   & 1.1 / 1.2                           & 1.9 / 2.8                           \\
            concat pairs   & \cellcolor{lightgray!50}24.7 / 29.6 & \cellcolor{lightgray!50}28.5 / 35.6 \\
            \bottomrule
        \end{tabularx}
    } \hfill
    \subfloat[Different numbers of relation queries.]{
        \label{tab:ablation_e}
        \begin{tabularx}{0.29\textwidth}{c|c c}
            \toprule[0.15em]
            \# Rel-Query & mR/R@20                             & mR/R@50                             \\
            \midrule[0.15em]
            50           & 23.4 / 26.7                         & 29.7 / 35.7                         \\
            100          & \cellcolor{lightgray!50}24.7 / 29.6 & \cellcolor{lightgray!50}28.5 / 35.6 \\
            200          & 22.3 / 29.7                         & 25.9 / 35.6                         \\
            \bottomrule
        \end{tabularx}
}
    
\end{table*}

\subsection{Ablation Study}

\vspace{2mm}
\noindent
\textbf{Ablation study on each component of PPN.} In \Cref{tab:ablation_a}, we first perform ablation studies on the effectiveness of each component of PPN. We find that all three components: \textit{linear embedding}, \textit{matrix learner}, and \textit{the supervision from directed adjacency matrix} are all important to the performance. Notably, without the linear embedding, the model will just diverge and not provide any correct prediction. This is because the object queries only contain the category information and ignore the pair-wised information. The embedding heads force the object queries to distinguish between subject and object. As shown in the second row of~\Cref{tab:ablation_a}, a similar situation also happens on the BCE supervision part, which provides vital information about the pair distributions that help pair proposal matrix learning. Adding Matrix Learner further improves the performance by filtering unconfident pairs, as shown in the third row of~\Cref{tab:ablation_a}.

\vspace{2mm}
\noindent
\textbf{Different Architectures for Matrix Learner.}
In~\Cref{tab:ablation_b}, we compare different types of architectures for the matrix learner. We select multi-layer perceptrons (MLP), two types of transformers, and two types of convolutional neural networks with model sizes between 0.2M and 30M. For the design of transformers, we follow the ViT~\cite{vit} by considering the $\mathbf{M}_{\text{rough}}$ as an image with the size of $\mathbb{R}^{1 \times N_\text{obj} \times N_\text{obj}}$. $\mathbf{M}_{\text{rough}}$ is projected by a patch embedding layer followed by three transformer layers with global or window attention then de-patchify it. Their results are represented as G-transformer and W-transformer in the Table respectively. The CNN-tiny represents a three-layer CNN with inner channel $64$ and kernel size $7$. In addition, we expand the CNN-tiny to CNN-base by two magnitudes (30M) for further comparison. Compared with MLP and transformer, CNN achieves the best results since the CNN-based matrix learner can filter out redundant noise and reserve the local details in the matrix, working as an efficient semantic filter. Moreover, increasing CNN parameters does not bring about major changes in the results, proving that the current scale of Matrix Learner is capable of filtering.

\vspace{2mm}
\noindent
\textbf{Ablation on Relation Loss.} To handle the long-tailed problems in~\Cref{sec:long-tail}, in~\Cref{tab:ablation_c}, we explore several balanced loss from the existing methods. We set the cross-entropy loss as the baseline. For Focal Loss~\cite{focalloss}, we test different settings $\gamma = 0, 0.5, 2$ and report the best one. As shown in that table, we choose the Seasaw Loss~\cite{seesawloss} for our relation classification loss.

\vspace{2mm}
\noindent
\textbf{Different Input for Key and Value of the Relation Fusion.} In~\Cref{tab:ablation_d}, we select four different levels of abstraction of information, from low to high, as the key and value input for the relation decoder. 
`Random initialization' means that the key and value vectors are both random initialized as trainable parameters. 
`Image features' means that we use the output from the vision encoder (ResNet-50\cite{resnet} in our case), as the KV inputs, which provides more information than two random initialization vectors but still has not enough prior for the classification for the relation. 
We also try `Random subject-object' pairs.
This design indicates these object queries $Q_\text{obj}$ have contextual information at the object level but do not have a pair-level structure. 
For our method, the concatenation of subject-object pair,  $Q_\text{pair}$, gives both object-level context and pair-level selection. As shown in~\Cref{tab:ablation_d}, an input with a more specific and richer context can boost the performance of the whole model.

\vspace{2mm}
\noindent
\textbf{Different Number of Relation Queries.} In~\Cref{tab:ablation_e}, we adjust the number of relation queries from 100 to 50 and 200. We find that the number of relation queries does not greatly affect the result, and our model is robust to the number of relation queries. We set the relation query number to 100 in our model.

\begin{table*}[!ht]
    \centering
    \caption{\textbf{More Pair-Net ablation experiments on PSG.} We report mean Recall and Recall with K=20, 50. Our settings are marked in \colorbox{lightgray!50}{gray}.}
    \subfloat[Different weights of loss function.]{
        \label{tab:ablation_f}
        \begin{tabularx}{0.44\textwidth}{c|c c}
            \toprule[0.15em]
        Loss weights ($\lambda_\text{o}$/  $\lambda_\text{s}$/  $\lambda_\text{r}$/  $\lambda_\text{pp}$) & mR/R@20 & mR/R@50\\
            \midrule[0.15em]
            4 / 4 / 2 / 5 & \cellcolor{lightgray!50}24.7 / 29.6 & \cellcolor{lightgray!50}28.5 / 35.6 \\
            4 / 4 / 2 / 10 & 22.0 / 25.7 & 26.8 / 32.4\\
            4 / 4 / 4 / 5 & 23.8 / 27.9 & 26.2 / 32.6\\
            8 / 8 / 2 / 5 & 22.2 / 26.5 & 26.7 / 32.9 \\
        \bottomrule[0.1em]
        \end{tabularx}
    }
    \hfill
    \subfloat[Effect of positive weight adjustment in BCELoss.]{
        \label{tab:ablation_g}
        \begin{tabularx}{0.44\textwidth}{c|c c}
            \toprule[0.15em]
        Positive weight adjustment & mR/R@20 & mR/R@50\\
        \midrule[0.15em]
        $\times$ & 0.6 / 1.2 & 1.2 / 2.3\\
        \checkmark & \cellcolor{lightgray!50}24.7 / 29.6 & \cellcolor{lightgray!50}28.5 / 35.6 \\
        \bottomrule[0.1em]
        \end{tabularx}
    } 
\end{table*}

\vspace{2mm}
\noindent
\textbf{Effect of different loss weights.} In~\Cref{tab:ablation_f}, we adjust the weights of components in the loss function. However, the different weights have no significant effects on the final results. This means that our model design is robust to different loss weight settings.

\vspace{2mm}
\noindent
\textbf{Positive Weight Adjustment in BCELoss.} The positive weight in the PPN is dynamically calculated by the ratio between the total size of $\mathbf{M}_{\text{gt}}$ and the number of positive samples in the $\mathbf{M}_\text{gt}$ hence it is not a hyperparameter. In~\Cref{tab:ablation_g}, we validate that performance dramatically decreases given the absence of the positive weight adjustment of the BCELoss. 

\subsection{Qualitative Results and Visualization} 
\begin{figure*}[t]
    \centering
    \includegraphics[width=0.90\linewidth]{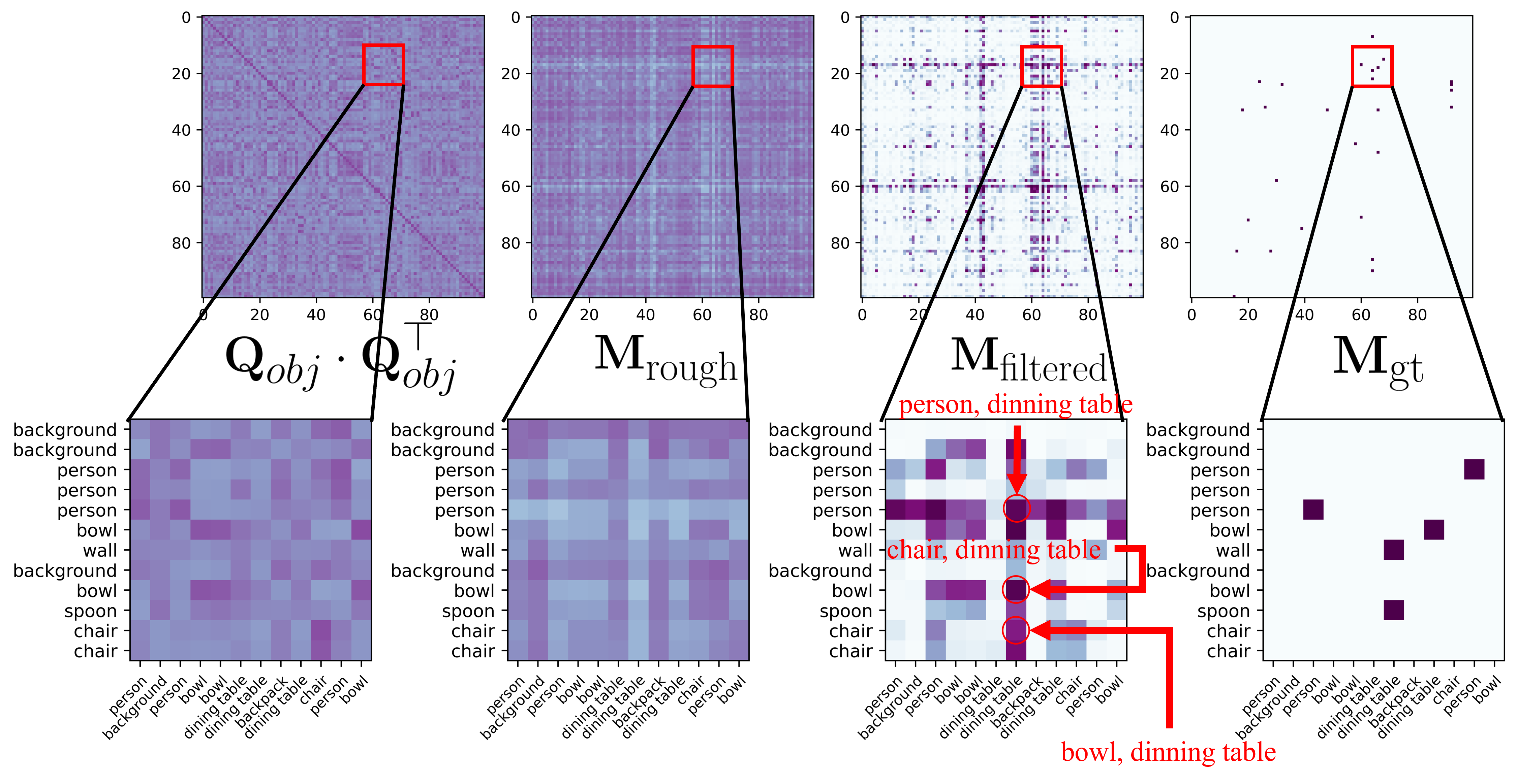}
    \caption{\textbf{The visualization of Pair Proposal Matrix.} Left to right: self-multiplication of object query $\mathbf{Q}_\text{obj}\cdot \mathbf{Q}_\text{obj}^{\top}$, $\mathbf{M}_{\text{rough}}$, $\mathbf{M}_{\text{filtered}}$, and $\mathbf{M}_{\text{gt}}$. It reflects that the pairing process in PPN is not based solely on semantic similarity and shows the necessity of the PPN.}
    \label{fig:ppn}
\end{figure*}

\textbf{Effect of Pair Proposal Network.} In \Cref{fig:ppn}, we visualize the functionality of each component in the PPN. From $\mathbf{Q}_{\text{obj}} \cdot \mathbf{Q}_{\text{obj}}^{\top}$ to $\mathbf{M}_{\text{rough}}$, the clear diagonal line is absent, showing that after the object/subject embedding projection, the value of the matrix is not \textit{solely} based on semantic similarity but starts to reflect the relational correlation between objects and subjects. Also, the matrix becomes non-symmetric, indicating that the projection correctly distinguishes the subject and object from the same inputs. After adopting our proposed Matrix Learner, from $\mathbf{M}_{\text{rough}}$ to $\mathbf{M}_{\text{filtered}}$, the matrix becomes sparse and clean. It is prone to better reflect relational correlation and depicts the capability of the Matrix Learner on filtering. Finally, the comparison between $\mathbf{M}_{\text{filtered}}$ and $\mathbf{M}_{\text{gt}}$ shows that although $\mathbf{M}_{\text{gt}}$ provides an elegant supervision signal to the network, it is unnecessary to reach exactly ground-truth sparsity, considering unlabeled but reasonable relations and possibly redundant object queries introduced in the annotation process.

\begin{figure*}[t]
    \centering
    \includegraphics[width=0.85\linewidth]{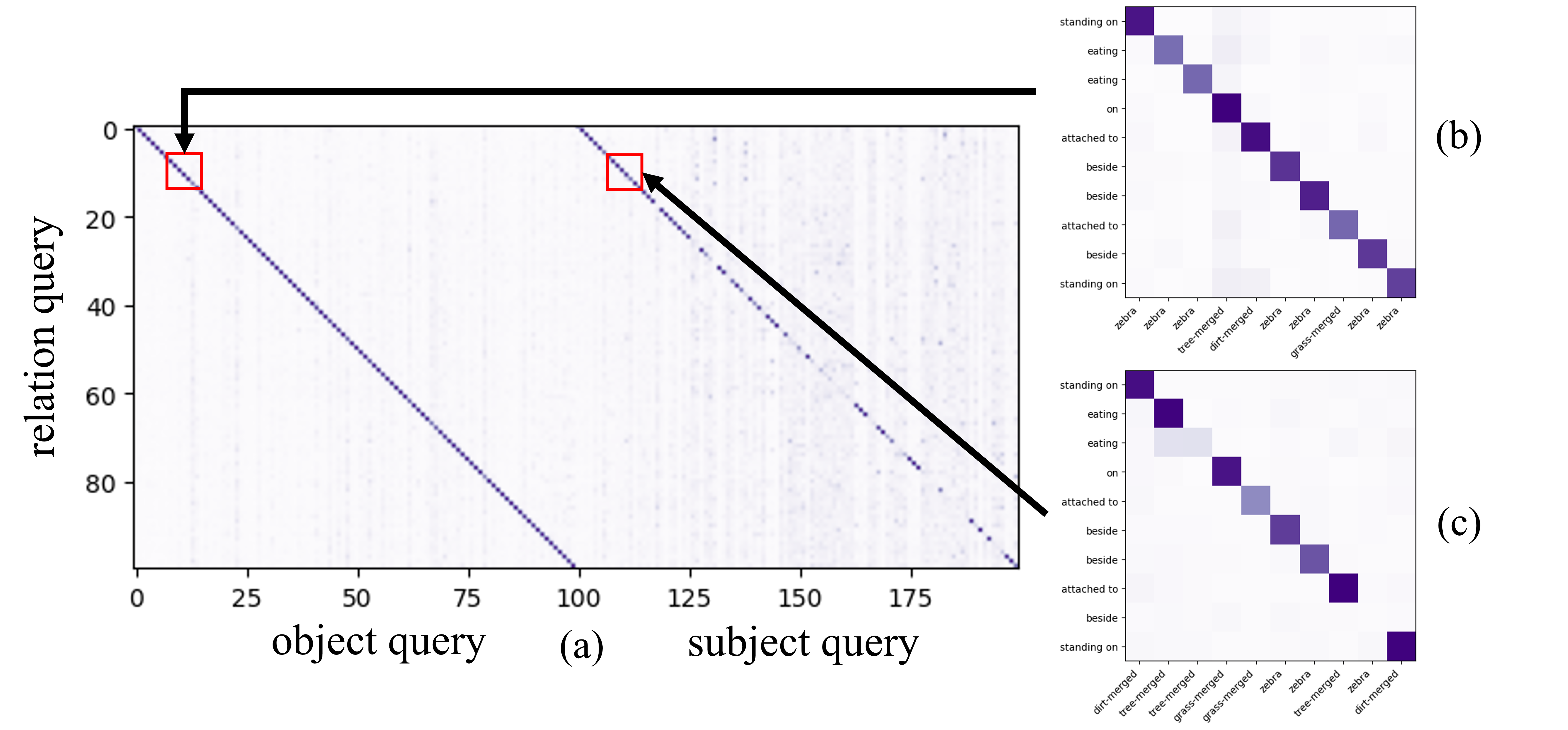}
    \caption{\textbf{Visualization of the cross-attention map of relation decoder with class annotations.} (a) is the overall $N_\text{rel} \times 2N_\text{rel}$ cross-attention map of the last layer of the relation decoder between the relation query $\mathbf{Q}_\text{rel}$ and the pair query $\mathbf{Q}_\text{pair}$. (b) and (c) are two selected detailed zones from (a) with class annotations of \texttt{\{Subject}, \texttt{Relation}, \texttt{Object\}}. The two diagonals in the figure show the strong correlation between the relation query and the matched subject/object query.}
    \label{fig:attn_map_vis}
\end{figure*}

\vspace{2mm}
\noindent
\textbf{Effect of Relation Fusion.} In~\Cref{fig:attn_map_vis}, we visualize the averaged cross-attention map of the last layer of the relation decoder. From (a), we notice from values of two diagonals that the $i$-th relation query is heavily weighted from the $i$-th subject query and the $i$-th object query, with minimal information from other queries. From (b) and (c), which shows a $10\times 10$ detailed region from (a) along the diagonals, we can see that the cross attention implicitly performs the matching process between pairs and relation to form the result triplet. This proves the relation fusion module successfully helps relation queries to find the matching subject query and object query.

\begin{figure*}[htbp]
    \centering
    \begin{subfigure}[b]{0.48\textwidth}
        \centering
        \includegraphics[width=\textwidth]{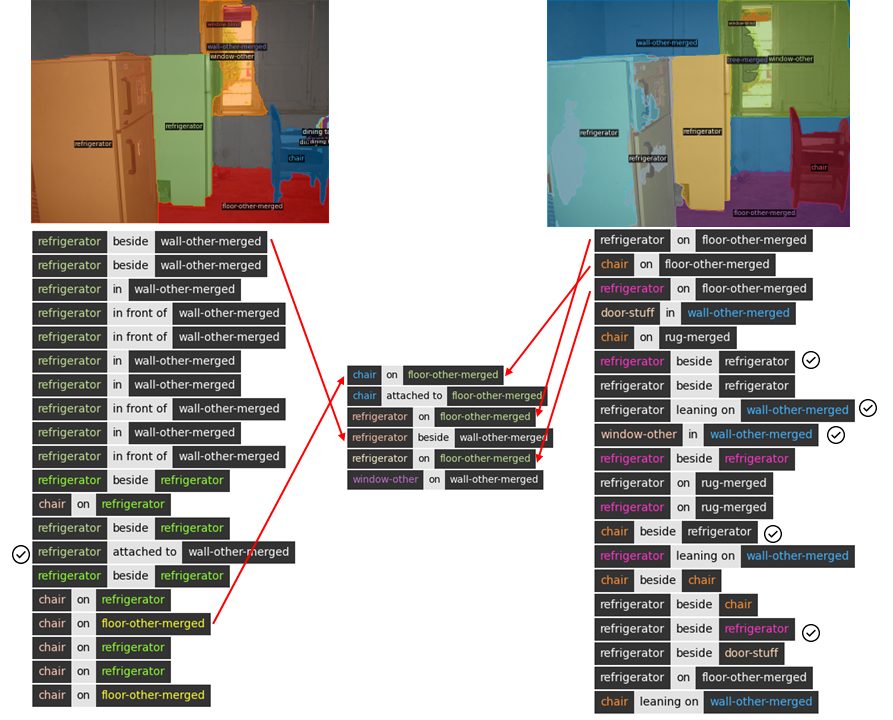}
        \label{fig:figure1}
    \end{subfigure}
    \hfill
    \begin{subfigure}[b]{0.5\textwidth}
        \centering
        \includegraphics[width=\textwidth]{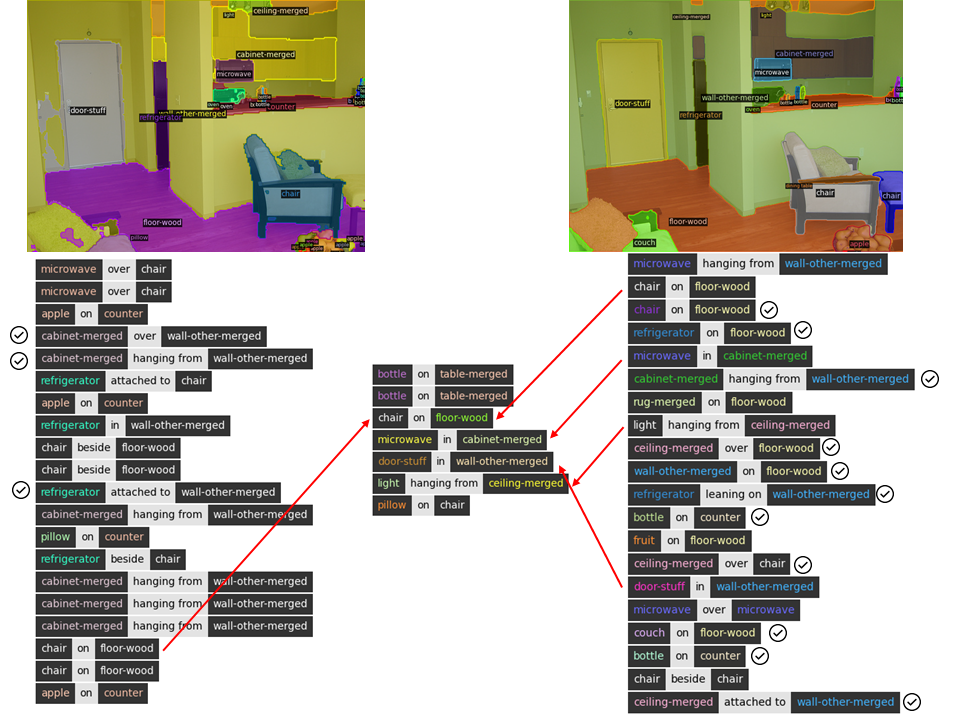}
        \label{fig:figure2}
    \end{subfigure}

    \begin{subfigure}[b]{0.48\textwidth}
        \centering
        \includegraphics[width=\textwidth]{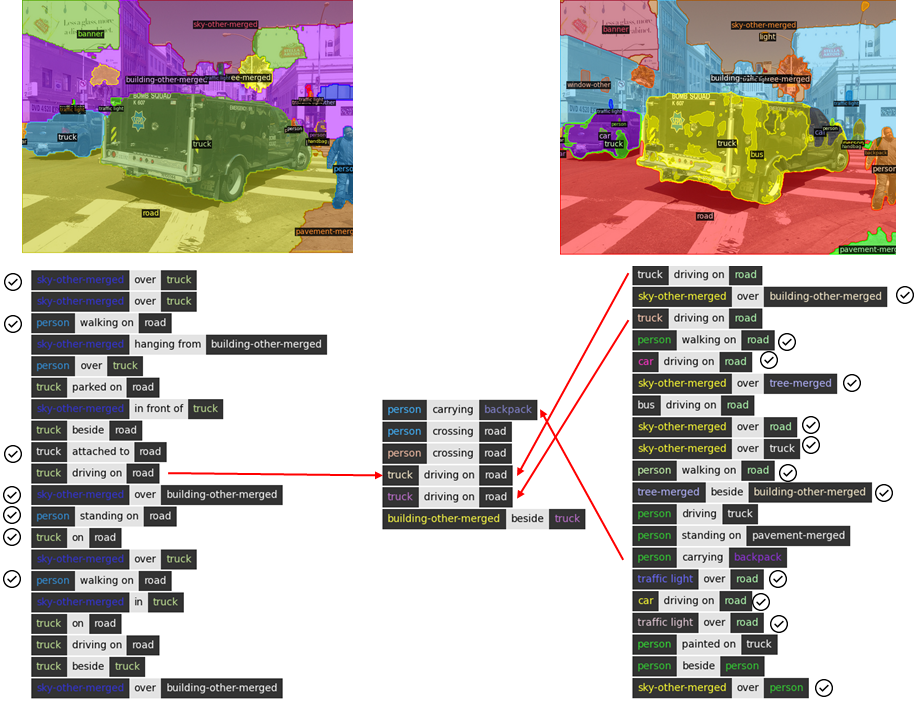} 
        \label{fig:figure3}
    \end{subfigure}
    \hfill
    \begin{subfigure}[b]{0.5\textwidth}
        \centering
        \includegraphics[width=\textwidth]{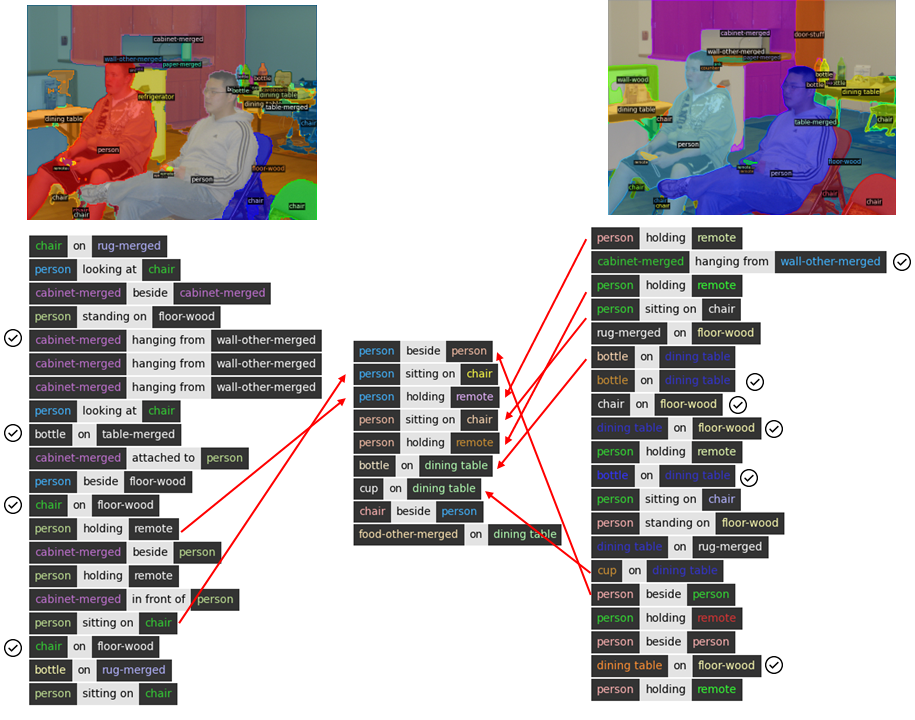} 
        \label{fig:figure4}
    \end{subfigure}

    \caption{\textbf{The visualization of scene graph generation of baseline model and Pair-Net.} In each image, the left column displays the results from the baseline model, the middle column shows the ground truth and the right column showcases the results from our proposed Pair-Net.}
    \label{fig:sg_vis}
\end{figure*}

\vspace{2mm}
\noindent
\textbf{Visualization of results on PSG.\label{sec:vis_sg_psg}}
\noindent
We further visualize the panoptic segmentation and top-20 scene graph triplet from our baseline, the Pair-Net, and the corresponding ground truths in \Cref{fig:sg_vis}. 
The center triplets are ground truth triplets, the left segmentation mask and triplets are provided by the baseline (PSGFormer) model, right image and triplets are from our Pair-Net. We use red arrows to correct predictions from the model and the ground truth. Meanwhile, we use a round tick mark beside the prediction to indicate the triplets which are not included in the ground truth annotation but are reasonable from a human's perspective. Such predictions should be considered useful information for the scene but do not reflect in the numerical evaluation metrics due to the missing annotations of the PSG dataset. We shall notice that the results from the baseline model produce many duplications of the same triplets. Such cases downgrade the performance of the model as there are fewer different triplets predictions given the top-k predictions,  which decrease the Recall@K and Mean Recall@K. Such duplication case is eliminated in the Pair-Net because of the pair-then-relation approach. In this process, we explicitly select a meaningful subject-object pair first, which eliminates the probability of producing duplicated triplets and encourages the model to produce harder and more diverse triplets from the scene.

\subsection{Experiments on VG-150\label{sec:exp_vg}}
\noindent
We further report the performance of Pair-Net on VG-150 \cite{vg} in \Cref{tab:main_vg} to show that our method could be generalized to a bounding-box-level scene graph dataset and achieves comparable performances with the current specific designed SGG models. 

\vspace{2mm}
\noindent
\textbf{Visual Genome (VG) \cite{vg}.} We use the most widely used variant of VG~\cite{vg80k}, namely VG-150, which includes 150 object classes and 50 relation classes. We mainly adopt the data splits and pre-processing from the previous works \cite{imp, motifs, bi-graph}. After filtering, the VG-150 contains 62k and 26k images for training and testing, respectively. 
\begin{table*}
    \centering
        \caption{\textbf{Results on VG-150 dataset.} Pair-Net achieves comparable performance in recall and mean recall compared with previous transformer-based SOTA works. '-' indicates that the original paper does not report the data.}
    \label{tab:main_vg}
    \begin{tabular}{c|c|c|ccc|ccc}
        \toprule[0.15em]
        Backbone & Detector & Model & mR@20 & mR@50 & mR@100 & R@20 & R@50 & R@100\\
        \midrule[0.15em]
        \multirow{3}*{VGG\cite{vgg}} & \multirow{5}*{Faster R-CNN\cite{faster_rcnn}} & IMP \cite{imp} & 2.8 & 4.2 & 5.3 & 18.1 & 25.9 & 31.2\\
        & & MOTIFS \cite{motifs}  & 4.1 & 5.5 & 6.8 & \textbf{25.1} & \textbf{32.1} & \textbf{36.9} \\ 
        & & VCTree \cite{vctree}  & 5.4 & 7.4 & 8.7 & 24.5 & 31.7 & 36.3 \\
        \cmidrule(){1-1} \cmidrule(){3-9}
        \multirow{3}*{ResNeXt-101~\cite{resnext}} & & RelDN \cite{contrastive-loss}  & - & 6.0 & 7.3 & 22.5 & 31.0 & 36.7 \\
        & & GPS-Net \cite{gpsnet} & - & 7.0 & 8.6  & 22.3 & 28.9 & 33.2 \\
        \cmidrule(){1-9}
        \multirow{2}*{ResNet-101~\cite{resnet}} & {DETR~\cite{detr}} & SGTR~\cite{sgtr} & - &  12.0 & 15.2  & - & 24.6 & 28.4 \\ 
        & {Deformable DETR~\cite{deformable_detr}} & \textbf{Pair-Net-Bbox (Ours)} &  \textbf{8.9} & \textbf{12.4} & \textbf{15.4} & 18.8 & 24.9 & 29.3 \\
        \bottomrule
    \end{tabular}
\end{table*}

\begin{table}
    \centering
    \caption{\textbf{The performance of object detectors on VG-150.} Different models have a similar object detection ability on VG-150 due to the low-quality bounding box label.}
    \label{tab:od_vg}
    \scalebox{0.90}{
    \begin{tabular}{c|c|cc}
        \toprule[0.15em]
        Backbone & Detector & Model & AP$_{50}$\\
        \toprule[0.15em]
        \multirow{2}*{ResNet-101~\cite{resnet}} & {DETR~\cite{detr}} & SGTR~\cite{sgtr} & 31.2 \\ 
        & {Deformable DETR~\cite{deformable_detr}} & \textbf{Pair-Net-Bbox (Ours)} & 31.2 \\
        \bottomrule[0.1em]
    \end{tabular}}
\end{table}

\vspace{2mm}
\noindent
\textbf{Training configuration.} We fine-tune a 100-query Deformable DETR \cite{deformable_detr} on VG-150 for the object detection task for 30 epochs. The rest training configurations are the same as training on PSG. 
To adapt this task, we replace the Mask2Former~\cite{mask2former} in Pair-Net with the Deformable DETR~\cite{deformable_detr} as the detector, dubbed as \textbf{Pair-Net-Bbox}. 
We also change the backbone from ResNet-50 to ResNet-101 for a fair comparison with previous work. The rest of the architecture is the same as Pair-Net.


\vspace{2mm}
\noindent
\textbf{Results on VG-150 Benchmark.} In \Cref{tab:main_vg}, we apply Recall@20/50/100 and Mean Recall@20/50/100 as our benchmarks. 
We mainly compare our results with the previous transformer-based VG-150 SOTA model: SGTR \cite{sgtr}. For recall, our methods gain a + 0.3 gain in R@50 and + 0.9 in R@100. 
%
Considering mean recall, our method performs comparably with an additional $0.2\sim 0.4$ improvement. 
It could be noted that our method is \textit{not} specially designed for SGG, and we do not perform any extra changes from PSG to SGG. We will put these as our future work.\\

\vspace{2mm}
\noindent
\textbf{Influence of Object Detector.} Shown in \Cref{tab:od_vg}, previous SGG models and ours have a similar ability in object detection in terms of mAP$_{50}$ on VG-150~\cite{vg}. 
Following the previous discussion~\cref{tab:pair_recall} that the panoptic segmentation performance in terms of PQ is not the key factor to the performance of PSG models, object detection ability in terms of $AP_{50}$ is also not the key factor to the performance in SGG task. The model's performance is mainly determined by the quality of the pair classification and relation classification and not bottleneck by the object detection ability.
\section{Conclusion}
\label{sec:conl}
In this work, to tackle the challenging PSG task, we first conduct an in-depth analysis and present valuable insights for PSG research. We highlight the importance of accurate subject-object pairing. Based on these insights, we propose Pair-Net, a simple and effective framework that achieves state-of-the-art performance on the PSG dataset. We design the Pair then Relation network (Pair-Net), which has a novel Matrix Learner to directly learn the sparse pair-wise relation among the object queries. We hope this work can help advance research in this field and provide a stronger baseline for PSG's downstream tasks.

\vspace{2mm}
\noindent
\textbf{Limitation and Societal Impacts.} One limitation of Pair-Net is that we only explore a middle-scale dataset, i.e., PSG. This setting is mainly for a fair comparison with other works~\cite{psg}. Exploring a larger SGG dataset \cite{oi} will be our future work. We hope that other CV domains, like Visual Grounding and Visual Question Answering, can gain some insights from the pair-wised relations.

\vspace{2mm}
\noindent
\textbf{ACKNOWLEDGMENTS.} This work is supported by the National Research Foundation, Singapore under its AI Singapore Programme (AISG Award No: AISG2-PhD-2021-08-019), NTU NAP, MOE
AcRF Tier 2 (T2EP20221-0012), and under the RIE2020 Industry Alignment Fund - Industry Collaboration Projects (IAF-ICP) Funding Initiative, as well as cash and in-kind contribution from the industry partner(s).

\ifCLASSOPTIONcaptionsoff
  \newpage
\fi

{
\bibliographystyle{IEEEtran}
\bibliography{IEEEabrv,egbib}
}

\end{document}